\documentclass[pdflatex,sn-mathphys-num]{sn-jnl}

\usepackage[utf8]{inputenc} % allow utf-8 input
\usepackage[T1]{fontenc}    % use 8-bit T1 fonts

\usepackage{graphicx}%
\usepackage{multirow}%
\usepackage{amsmath,amssymb,amsfonts}%
\usepackage{amsthm}%
\usepackage{mathrsfs}%
\usepackage[title]{appendix}%
\usepackage{xcolor}%
\usepackage{textcomp}%
\usepackage{manyfoot}%
\usepackage{algorithm}%
\usepackage{algorithmicx}%
\usepackage{algpseudocode}%
\usepackage{listings}%

\usepackage{hyperref}       % hyperlinks
\usepackage{url}            % simple URL typesetting
\usepackage{booktabs}       % professional-quality tables
\usepackage{nicefrac}       % compact symbols for 1/2, etc.
\usepackage{microtype}      % microtypography
\usepackage{mathtools}

\usepackage{tcolorbox} % box to highlight the main message

\usepackage[capitalize]{cleveref}

\newcommand{\colorline}[1]{\hspace{-0.01\linewidth}\colorbox{green!30}{#1}}
\newcommand{\colortext}[1]{\colorbox{green!30}{#1}}

\crefname{section}{Sec.}{Secs.}

\begin{document}

\title{Initialisation and Network Effects in Decentralised Federated Learning}

\author*[1,2]{\fnm{Arash} \sur{Badie-Modiri}}\email{academic@arash.network}

\author[3]{\fnm{Chiara} \sur{Boldrini}}\email{chiara.boldrini@iit.cnr.it}

\author[3]{\fnm{Lorenzo} \sur{Valerio}}\email{lorenzo.valerio@iit.cnr.it}

\author[1,4]{\fnm{János} \sur{Kertész}}\email{kerteszj@ceu.edu}

\author[1]{\fnm{Márton} \sur{Karsai}}\email{karsaim@ceu.edu}

\affil*[1]{\orgdiv{Department of Network and Data Science}, \orgname{Central European University}, \orgaddress{\postcode{1100} \city{Vienna}, \country{Austria}}}

\affil[2]{\orgdiv{School of Science}, \orgname{Aalto University}, \orgaddress{\postcode{02150} \city{Espoo}, \country{Finland}}}

\affil[3]{\orgdiv{The Institute of Informatics and Telematics}, \orgname{National Research Council}, \orgaddress{\postcode{56124} \city{Pisa}, \country{Italy}}}

\affil[4]{\orgdiv{\orgname{Complexity Science Hub}, \orgaddress{\postcode{1080} \city{Vienna}, \country{Austria}}}}

\abstract{Fully decentralised federated learning enables collaborative training of individual machine learning models on a distributed network of communicating devices while keeping the training data localised on each node. This approach avoids central coordination, enhances data privacy and eliminates the risk of a single point of failure. Our research highlights that the effectiveness of decentralised federated learning is significantly influenced by the network topology of connected devices and the initial conditions of the learning models. We propose a strategy for uncoordinated initialisation of the artificial neural networks based on the distribution of eigenvector centralities of the underlying communication network, leading to a radically improved training efficiency. Additionally, our study explores the scaling behaviour and the choice of environmental parameters under our proposed initialisation strategy. This work paves the way for more efficient and scalable artificial neural network training in a distributed and uncoordinated environment, offering a deeper understanding of the intertwining roles of network structure and learning dynamics.}


\maketitle

% TLDR: We propose a model parameter initialisation strategy for uncoordinated decentralised federated learning based on the distribution of centralities of the communication network, and show the effects of topology of that network on the training curve.

\section{Introduction}\label{sec:introduction}
The traditional centralised approach to machine learning has shown great progress in the last few decades. This approach, while practical, comes at a cost in terms of systemic data privacy risks and centralisation overhead~\cite{mcmahan2017communication, kairouz2021advances, rieke2020future}. To alleviate these issues, the \emph{federated learning} framework was proposed where each node (client) updates a local machine learning model using local data and only shares its model parameters with a centralised server, which in turn aggregates these individual models into one model and redistributes it to each node~\cite{mcmahan2017communication}.

While this approach reduces the data privacy risk by eliminating data sharing with the centralised server, it still maintains a singular point of failure and puts a heavy communication burden on the central server node~\cite{beltran2023decentralized, lalitha2018fully}. \emph{Decentralised federated learning} aims to provide an alternative approach that maintains data privacy but removes the need for a centralised server. This involves the set of nodes (clients) updating their local models based on the local data, but directly communicating with one another through a communication network. Each node then updates its local model by aggregating those of the neighbourhood~\cite{beltran2023decentralized, valerio2023coordination}.

The efficiency of this approach is heavily impacted by several kinds of inhomogeneities~\cite{valerio2023coordination} characterising the network structure, initial conditions, distribution of learning data, and temporal irregularities. In this paper, we focus on the effect of network structure and the initialisation of parameters, showing how the system in its early stages shows diffusion dynamics on networks and how this understanding helps us design an optimal fully-decentralised initialisation algorithm based on eigenvector centralities of the connectivity network.

\paragraph{Motivation}
Decentralised federated learning immediately raises two distinct new issues compared to the centralised federated learning approach. First, \textbf{that the initialisation and operations of the nodes have to be performed in an uncoordinated manner}, as the role of coordination previously lay with the server.

Second, \textbf{that the effect of the structure and heterogeneities in the communication network is poorly understood}. In the case of centralised federated learning, the communication network is organised as a simple star graph (\cref{fig:coordinated-uncoordinated}a). In a decentralised setting (\cref{fig:coordinated-uncoordinated}b-c), however, the network might be an emergent result of, e.g., the social network of the users of the devices, a distributed peer-discovery protocol or a hand-engineered topology comprised of IoT devices. Each of these assumptions leads to a different network topology with wildly different characteristics. Many network topologies modelling real-world phenomena, unlike a star graph, have diameters that scale up with the number of nodes, inducing an inherent latency in the communication of information between nodes that are not directly connected. Structural heterogeneities, e.g., the dimensionality of the network, degree heterogeneity and heterogeneities in other centrality measures also play important roles in the evolution of various information-sharing dynamics on networks~\cite{newman2010Book}. This makes network heterogeneities primary candidates for analysis of any decentralised system.

\paragraph{Contribution}\label{sec:contribution}

Our research investigates the interplay between local model initialization and the underlying communication network topology in decentralised learning scenarios. Our contributions are as follows:
\begin{itemize}
\item We demonstrate that conventional model initialization methods, widely adopted in centralised settings, yield suboptimal performance when training occurs collaboratively in decentralised environments with complex communication topologies.
\item We introduce a theoretically grounded, novel uncoordinated initialization strategy that effectively addresses this limitation, significantly outperforming standard initialization techniques originally tailored for centralised or coordinated scenarios.
\item We empirically validate the proposed initialization method across a diverse set of network topologies, confirming its robustness and superior performance independent of the specific topology used.
\item Additionally, we analyse and characterize how different aspects of communication network topology influence the scaling properties and overall efficiency of the decentralised learning process.
\end{itemize}

\section{Related works}
Decentralised federated learning~\cite{beltran2023decentralized} comes as a natural next step in the development of the field of federated learning to save communication costs, improve robustness and privacy~\cite{mcmahan2017communication}. This approach has been used in application areas such as object recognition in medical images~\cite{roy2019braintorrent, tedeschini2022decentralized} and other industrial settings~\cite{savazzi2021opportunities, qu2020blockchained}. It has also been extended by novel optimisation and aggregation methods~\cite{sun2022decentralized, lalitha2018fully, lian2017can} and theoretical advances in terms of convergence analysis \cite{koloskova2020unified}.

The structure of complex networks, central to decentralised federated learning by coding the communication structure between connected devices~\cite{yuan2024survey}, can embody various heterogeneities. These have been found to be a crucial factor in understanding a variety of \emph{complex systems} that involve many entities communicating or interacting together. For example, the role of degree distribution~\cite{jennings1937structure, albert2000error}, high clustering~\cite{watts1998collective, luce1949method} or the existence of flat or hierarchical community structures~\cite{rice1927identification, fortunato2016community} in networks of real-world phenomena has been understood and analysed for decades. Recent advances in network modelling have extended heterogeneities in networks from structural to incorporate spatial~\cite{orsini2015quantifying} and also temporal heterogeneities, induced by patterns of, e.g., spatial constraints or bursty or self-exciting activity of the nodes~\cite{karsai2011small, gauvin2022randomized, badie2022directed_long, badie2022directed_short}. In the decentralised, federated learning settings, the structural heterogeneities of the underlying communication network have only been very recently subjected to systemic studies. Notably, \citet{vogels2022beyond} analyses the effect of topology on optimal learning rate and \citet{palmieri2023effect} analyses the differences among individual training curves of specific nodes (e.g., high-degree hubs versus peripheries) for Barabási–Albert networks and stochastic block models with two blocks~\cite{newman2010Book}.

On the matter of parameter initialisation in federated learning, recent studies have focused on the effect of starting all nodes from a homogeneous set of parameters~\cite{valerio2023coordination} or the parameters of an independently pre-trained model~\cite{nguyen2022begin}. Historically, artificial neural networks were initialised from random uniform or Gaussian distribution with scales set based on heuristics and trial and error~\cite{goodfellow2016deep} or for specific activation functions~\cite{lecun2002efficient}. The advent of much deeper architectures and widespread use of non-linear activation functions such as \textit{ReLU} or \textit{Tanh} led to a methodical understanding of the role of initial parameters to avoid exploding or diminishing activations and gradients. \citet{glorot2010understanding} proposed a method based on certain simplifying assumptions about the non-linearities used. Later, \citet{he2015delving} defined a more general framework for use with a wider variety of non-linearities, which was used for training the \textit{ResNet} image recognition architecture~\cite{he2016deep}. 

While machine learning and deep learning in particular have seen significant progress in the last decade, there have been no significant developments in general neural network initialisation since~\cite{he2016deep}. To the extent that the term is meaningfully defined, the He initialization method can be considered state-of-the-art for initialisation of generalised neural network architectures. However, as we show in the paper, this method, on its own, might not be optimal for fully decentralised learning. Alternatively, there have been suggestions for distributing some amount of information to every node, for example a pre-trained model or some seed value whence an initial set of parameters can be drawn \cite{nguyen2022begin}. In this manuscript we solely focus on the more general case of \emph{fully decentralised federated learning}, where we cannot safely assume that any amount of information can be reliably distributed to every node.

In this work, we will be proposing an extension of the previous works on effective artificial neural network parameter initialisation to the decentralised setting, where the parameters are affected not only by the optimisation based on the training data but also by interactions with other nodes of the communication network. The outcome and the techniques introduced in this work ensure that the consecutive aggregations of parameters do not unduly compress artificial neural network parameters.

\section{Preliminaries}\label{sec:preliminaries}

\subsection{System model and notation}\label{sec:notation}
In our setup, we used a simple decentralised federated learning system with an iterative process.
Nodes are connected through a predetermined static, undirected communication network $G = (\mathcal{V}, \mathcal{E})$ where $\mathcal{E} \subseteq \{\{v_i, v_j\} | v_i, v_j \in \mathcal{V} \text{ and } v_i \neq v_j\}$. 
All nodes train the same artificial neural network architecture $h(w_i,\cdot)$ with $i$ indicating the parameters of the $i$-th node. Moreover, at time zero, we assume that $w_i \neq w_j \forall i,j \in \mathcal{V}$.
$n = |\mathcal{V}|$ indicates the number of nodes (sometimes referred to as the system size), while $k_i = |\mathcal{N}_i|$ indicates the degree of a node $i$ defined as the number of its neighbours (where $\mathcal{N}_i$ denotes the set of neighbours for node $i$), and $p(k)$ is the degree distribution. For a node reached by following a random link, $q(k)$ is the distribution of the number of other links to that node, known as the excess degree distribution. The adjacency matrix is indicated by $A$.

Nodes $v_i \in \mathcal{V}$ initialise their local model parameters based on one of the following strategies: (1) random initialisation with no gain correction, an uncoordinated approach where each node draws their initial parameters independently based on a strategy optimised for isolated centralised training, e.g., from Ref.~\cite{he2015delving}; or (2) random initialisation with gain correction, which is our proposed initialisation strategy that re-scales initial parameter distributions from (1) based on the topology of the communication network. This approach will be explored in detail in \cref{sec:initialisation}.

Briefly, the initialisation method presented in \citet{he2015delving} sets the
weight variance $\sigma^2_l$ so that the variance of the pre-activation
$z^{(l)}$ in layer $l$ matches that of layer $l-1$. This keeps both forward activations and back-propagated gradients from exploding or vanishing at the start of training. For a layer with \emph{fan-in} $n_{\text{in}}$ (i.e., the number of inputs feeding each neuron) and element-wise activation $f$, the recommended weight variance is
\begin{equation}
  \sigma^2_l = \frac{g^{2}}{n_{\text{in}}},
  \qquad
  g^{2} = \frac{1}{\mathbb{E}_{z\sim\mathcal N(0,1)}[f'(z)^{2}]}\,,
\end{equation}
where the constant factor $g$ depends solely on the activation function $f$. For example, $g^{2}=2$ for ReLU, $g^{2}=1$ for linear or tanh activation and $g^{2}=2/(1+\alpha^{2})$ for leaky ReLU with negative slope $\alpha$ \cite{he2015delving}. Then, \citet{he2015delving} initialise the weights of layer $l$ using a Gaussian (or sometimes uniform) distribution with zero mean and variance $\sigma^2_l$.

Each node $i$ of the communication network holds a labelled local dataset $D_i$ such that $D_i \cap D_j = \emptyset, \forall i,j \in \mathcal{V}$ which are a portion of the same global dataset $D = \bigcup_{i=1}^{|\mathcal{V}|} D_i$. Each node can access its private data but not those of other nodes and we assume that the local datasets do not change over time. All nodes train the same artificial neural network architecture $h(w_i,\cdot)$ with $w_i \in \mathbb{R}^p$ indicating the parameters of the $i$-th node solving the following Empirical Risk Minimisation problem:
\begin{eqnarray}
    \min_{w} F(w_i;D_i), &  F(w_i;D_i)=\frac{1}{|D_i|}\sum_{(x,y)\in D_i}\ell(y,h(w_i;x))
\end{eqnarray}
with $\ell$ representing a generic loss function, which can be convex or non-convex. Moreover, at time zero we assume that $w_i \neq w_j, \forall i,j \in \mathcal{V}$. 
The nodes perform one or more batches of local training on their local data using an optimiser. At each iteration $t$, called here a \emph{communication round}, each node $i$ updates its parameters (weights and biases) using parameter values communicated by the nodes in its neighbourhood. This process is called \emph{aggregation}. In centralised federated learning, the simplest and most popular aggregation strategy is  \textsc{FedAvg}~\cite{mcmahan2017communication}, whereby models are aggregated according to a weighted average of their parameters. We assume that aggregation is performed in decentralised settings through the decentralised version of \textsc{FedAvg} (here called \textsc{DecAvg}):
\begin{equation}\label{eq:param-update}
w_i =  \beta_i w_i + \sum_{\forall j \in \mathcal{N}_i} \beta_j w_j,\quad \beta_i = \frac{|D_i|}{|D_i|+\sum_{\forall j \in \mathcal{N}_i} |D_j|}, \forall i \in \mathcal{V}\,.
\end{equation}
Since both the iid and non-iid data distributions used in the manuscript provide on expectation the same number of total items per node, we can assume $\beta_i \approx 1/(k_i+1)$.

The centralised federated learning using the simple parameter averaging aggregation method (\textsc{FedAvg}) can be viewed as a special case of the decentralised federated learning using the simple averaging aggregation (\textsc{DecAvg}) on a fully connected network, as at each step each node concurrently plays the role of the central server, setting its parameters to the average parameters of all other nodes. This means that to the extent that the results presented in this manuscript apply to fully connected networks, they can also be utilised to understand the behaviour of this configuration of a centralised federated learning process.

\begin{figure}
    \centering
    \includegraphics[width=\linewidth,trim={0 0 2cm 0},clip]{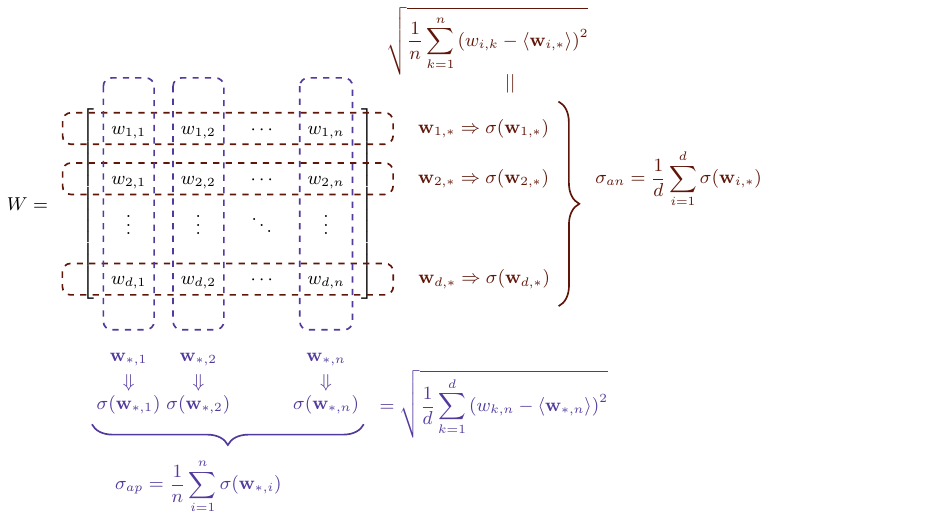}
    \caption{Illustration of matrix $W$, $w_{*,i}$, $w_{j,*}$, $\sigma_{ap}$ and $\sigma_{an}$. Each column of the matrix $W$ represents parameters of a single node, and each row represents the values of the same parameter on different nodes. $\sigma_{an}$ and $\sigma_{ap}$ can therefore be defined as the mean of the standard deviations of rows and columns of $W$, respectively.}
    \label{fig:ap_an}
\end{figure}

Expected values are indicated using the brackets around the random variable, e.g., $\langle k \rangle$ is the mean degree. Standard deviation is shown using $\sigma(\ldots)$. The $d$ learning model parameters of node $i$ are indicated by vector $w_i$ of size $d$, sometimes arranged in a $d \times n$ matrix $W = \{w_{j,i}\}$. $\sigma_{ap}$ indicates the mean standard deviation across columns of $W$, i.e.~$\sum_{i=1}^{n} \sigma(w_{*, i})/n$ where $w_{*, i}$ denotes column $i$ of $W$, while $\sigma_{an}$ indicates the mean standard deviation across rows, i.e.~$\sum_{j=1}^{d} \sigma(w_{j, *})/d$, where $w_{j, *}$ denotes row $j$ of $W$. In plain terms, $\sigma_{ap}$ represents the variability of the entire set of learning model parameters within each node, while $\sigma_{an}$ captures how each parameter varies across different nodes. An illustration of the definition of the notation is provided in \cref{fig:ap_an}.

It is important to note that artificial neural networks are usually initialised with parameters that are drawn from different sets of distributions, e.g., the weights of each layer can be drawn from a separate distribution with standard deviation selected based on the outgoing ``fan-out'' connections to the next layer. In this case, a vector $w_i = w_{*,i}$ can be formed for one specific set of parameters, drawn from a zero-mean distribution with standard deviation $\sigma_{init.}$.

\subsection{Experimental setup}\label{sec:setup}

To evaluate and demonstrate the effectiveness of the initialisation algorithm proposed in \cref{sec:initialisation}, as well as the scaling properties of the decentralised learning process in \cref{apdx:topology}, we rely on a variety of real-world datasets, neural network architectures and optimisers.

Our experiments will be performed on subsets of the MNIST digit classification task~\cite{lecun1998gradient}, the So2Sat LCZ42 dataset for local climate zone classification~\cite{zhu2019so2sat} and the CIFAR-10 image classification dataset~\cite{krizhevsky2009learning}. The data is distributed between nodes either \emph{iid}, i.e., each node receiving roughly an equal number of each class, or \emph{non-iid} based on a Zipf distribution, i.e., each node receiving roughly an equal number of total training samples, but they are not uniformly distributed across different classes.

The neural network architectures used in this work are (1) a simple multilayer perceptron (\emph{MLP}) consisting of three fully-connected hidden layers, (2) a simple convolutional neural network architecture (\emph{CNN}) consisting of 3 2D convolutional layers with 32, 64 and 64 output channels, each with 3 kernels and one pixel padding of zeros and (3) the more realistic \emph{VGG16} architecture~\cite{simonyan2014very} representing more modern, deeper architectures. The details of the experimental setup, as well as the runtimes for each experiment can be found in \cref{apdx:data-and-experiments}. Note that in the organisation of this manuscript, we intersperse analytical reasoning and methodological descriptions with empirical evaluation.

\section{Uncoordinated initialisation of artificial neural networks}\label{sec:initialisation}
\begin{figure}
    \centering
    \includegraphics[width=\linewidth]{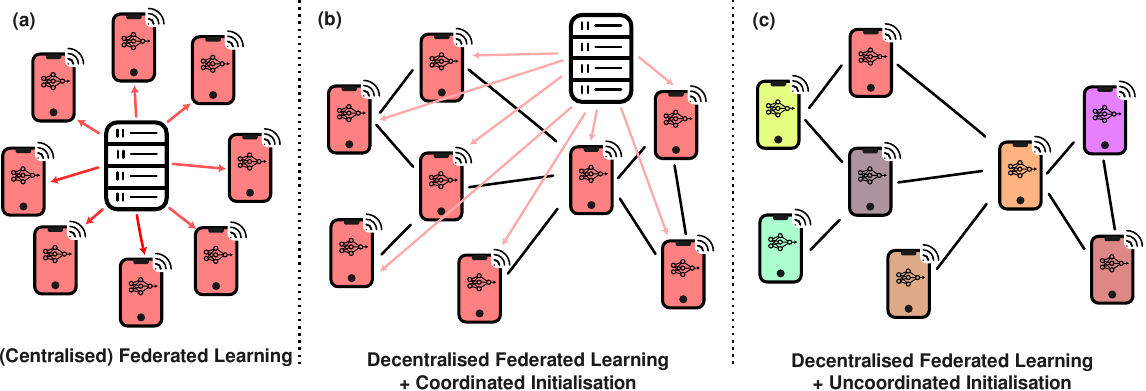}
    \caption{A comparison between (a) a typical centralised federated learning setup where nodes communicate only through a central server, (b) a typical decentralised but coordinated federated learning setup where nodes communicate directly with their peers, but a central server still plays a role in coordinating the setup and (c) a fully uncoordinated decentralised federated learning setup, where no coordination through a centralised server is necessary. Of particular interest to this work is the initialisation of learning model parameters, displayed using colours in this schematic. In centralised and coordinated cases, the coordinating server can ensure every node receives the same set of initial parameters, shown here using the colour of each node, while in the fully uncoordinated setting, we cannot make such assumptions.}
    \label{fig:coordinated-uncoordinated}
\end{figure}

Unlike in centralised federated learning, it is unwarranted to assume that in a massive decentralised federated learning setup, all nodes can negotiate and agree on initial values for model parameters in a manner that each parameter is exactly equal across all nodes. A fully uncoordinated method for selecting initial values for the model parameters means that each node should be able to draw initial values for its model parameters independently, with few or no communication in a manner that is robust to possible communication errors. A schematic visualisation of centralised, decentralised and uncoordinated decentralised federated learning setups is presented in \cref{fig:coordinated-uncoordinated}.

It has been shown that judicious choice of initialisation strategy can enable training of much deeper artificial neural networks~\cite{glorot2010understanding, he2015delving, he2016deep}. Specifically, a good parameter initialisation method leads to initial parameters that neither increase nor decrease activation values for consecutive layers exponentially~\cite{he2015delving, glorot2010understanding}. In the case of decentralised federated learning, this proves more challenging, as the aggregation step changes the distribution of parameters, meaning that \emph{the optimal initial value distributions are not only a function of the machine learning model architecture, but also affected by the communication network structure}.

\subsection{Motivating example}
\begin{figure}[!bt]
    \centering
    \includegraphics[width=\linewidth]{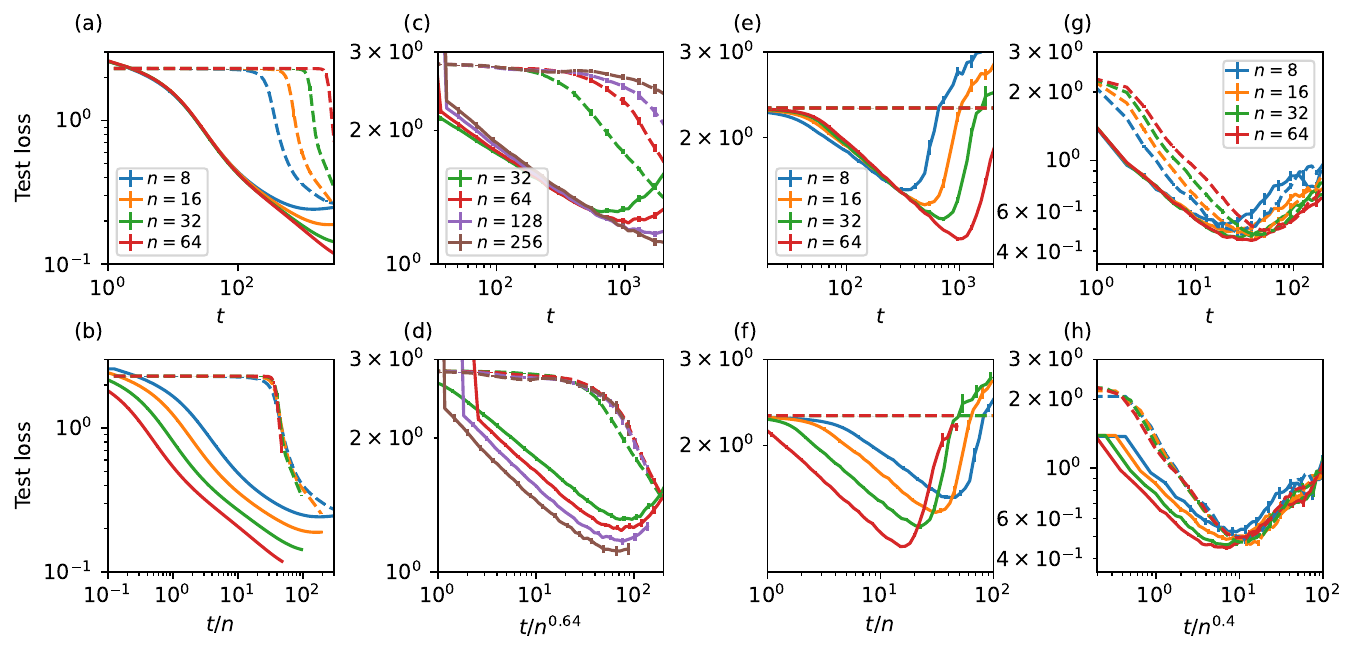}
    \caption{Mean test cross-entropy loss with the proposed initialisation (solid lines) compared to the initialisation method proposed in \citet{he2015delving} without re-scaling (dashed lines). The decentralised federated learning process on nodes connected through (a,b) fully-connected (complete) networks with MNIST classification task on a simple multilayer perceptron with iid data distribution (c,d) Barabási--Albert networks with average degree 4, with the So2Sat LCZ42 classification task, using a simple convolutional architecture, Zipf data distribution, $\alpha=1.8$, (e,f) random 4-regular networks with CIFAR-10 classification task with VGG16 architecture and (g,h) same configuration as (a,b) but using Adam  optimiser with decoupled weight decay. The results show that without the proposed re-scaling of the parameters, the mean test loss has a plateau that lasts a number of rounds proportional to (or sub-linear in) the system size. Bottom row (b,d,f,h) shows the empirical scaling of the test loss time trajectory of the independent~\cite{he2015delving} method initialisation with system size, with exponents ranging from 0.4 to 1. Error bars represent 95\% confidence intervals.}
    \label{fig:simulation_plateau}
\end{figure}

\begin{figure}
    \centering
    \includegraphics[width=0.6\linewidth]{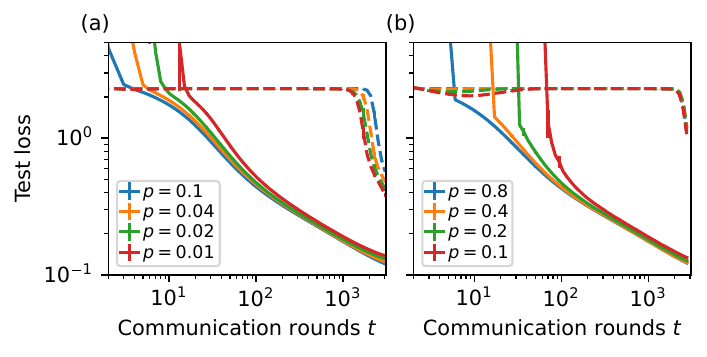}
    \caption{Mean cross-entropy test loss as a function of communication rounds for a fully connected communication network $n=64$ using the MNIST dataset with 512 items per node. Each (a) connection or (b) node is active at each round with probability $p$. Note that inactive nodes still perform local training, but are in effect momentarily isolated from the network. The proposed initialisation is displayed with solid lines and the independent initialisation method of \citet{he2015delving} with dashed lines. Even at fairly low values of $p$, the system as a whole has a much better overall learning trajectory with our proposed parameter initialisation method compared to that of \citet{he2015delving}. Error bars represent 95\% confidence intervals.}
    \label{fig:occupation}
\end{figure}

To illustrate the issues arising from blindly applying methods originally designed for centralised or coordinated learning settings to decentralised environments, we conducted extensive simulations of the standard DecAvg algorithm (configured as described in Section~\ref{sec:setup}), utilizing the initialization approach proposed by \citet{he2015delving}, which is widely recognized as the de-facto standard for centralised models. For benchmarking purposes, we compared the performance of independent initialization using the method of \citet{he2015delving} against our novel approach, detailed in Section~\ref{sec:sigma_ap} and represented by continuous lines in \cref{fig:simulation_plateau,fig:occupation}.

Empirically, we observe (\cref{fig:simulation_plateau} dashed lines) that the initialization strategy proposed by \citet{he2015delving} consistently leads to deteriorating performance (evidenced by delayed test loss reduction) in decentralised scenarios, particularly as the number of nodes increases. This observation holds true across various network topologies (complete, Barabási–Albert, and random 4-regular graphs), diverse model architectures (MLP, CNN, VGG16), data distributions (iid and Zipf), and optimization methods. \cref{fig:simulation_plateau}(b,d,f,h) shows this as scaling of the loss trajectory as a function of communication rounds with the number of nodes as $n^\mu$, with values of $\mu$ observed in range $0.4 \leq \mu \leq 1$ depending on the experimental setting.

In more realistic settings, it is often the case that, either due to faults, technical limitations or deliberate choice, not all communication channels between nodes stay open at all times, or that not all nodes participate at every round of communication. We analyse this by assuming each connection or node is active at each point in time with a probability $p$. \cref{fig:occupation} reveals that, under these conditions, the initialization method proposed by \citet{he2015delving} consistently exhibits degraded performance, even at relatively low activation probabilities $p$.
% In \cref{fig:occupation} we observe that our initialisation method performs favourably compared to the initialisation method of \citet{he2015delving} even at very low values of occupation probability $p$, with each node arriving at an eventually consistent state much earlier.

\subsection{Numerical model of early-stage dynamics}
\begin{figure}[!tb]
    \centering
    \includegraphics[width=\linewidth]{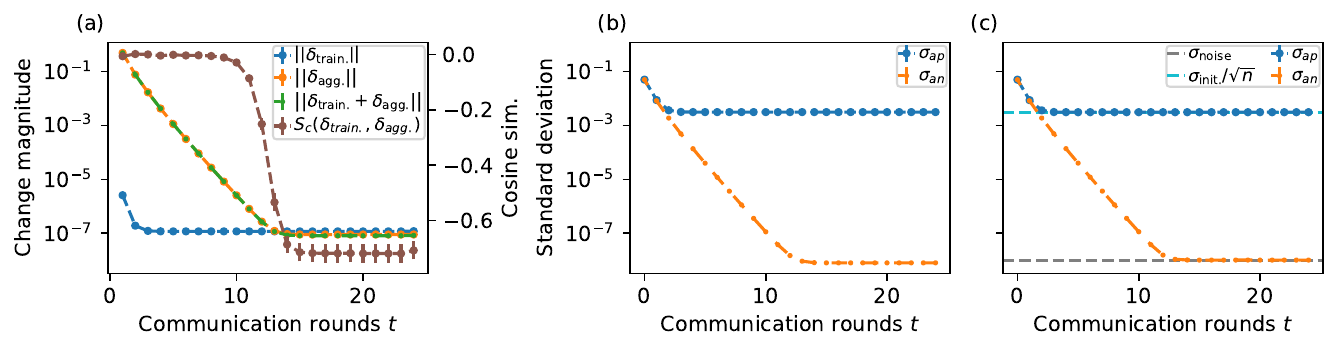}
    \caption{(a) Mean magnitude of change in parameters due to training and aggregation independently as well as the total change, as well as the mean cosine similarity of the changes during training and aggregation. In the early rounds of the iterative process, the vector of change due to the aggregation is several orders of magnitude larger than that of the training. Additionally, the cosine similarity trajectory indicates the orthogonality of these vectors in the early rounds, supporting the numeric model assumption that the early evolution of the system is dominated by the aggregation step. Additionally, the evolution of standard deviation of $\sigma_{an}$ and $\sigma_{ap}$ on (b) the distributed learning process with actual ANNs and (c) the numerical simplified model shows similar early-stage dynamics. Values were calculated by (a,b) running or (c) numerically modelling the decentralised federated learning process on random 32-regular $n=256$ networks. Panels (a,b) were performed with 80 training samples per node, 1 epoch per communication round. Error bars represent 95\% confidence intervals.}
    \label{fig:simulation_diffs}
\end{figure}

\begin{figure}
    \centering
    \includegraphics[width=\linewidth]{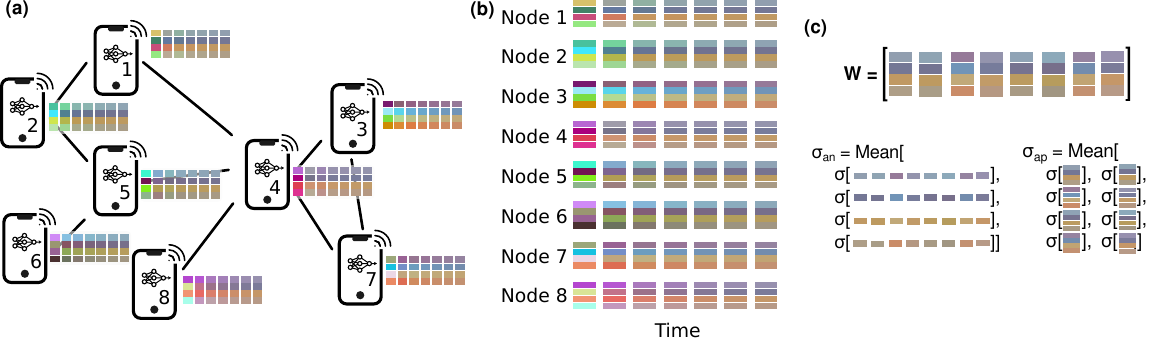}
    \caption{Simple visualisation of the parameter compression effect due to the aggregation process. Panel (a) shows the communication network, panel (b) shows the internal state of each node across time, with each parameter visualised as a colour and panel (c) shows the definition of parameter $W$, $\sigma_{an}$ and $\sigma_{ap}$ in this setting. In this visualisation, each node is assigned four independent parameters, shown using colours. As time progresses, the averaging process results in two outcomes: (1) the value of the same parameter across nodes converges toward each other, meaning that the average standard deviation of each parameter across nodes $\sigma_{an}$ decreases across time. This can be seen by the fact that, e.g., the third parameter across all nodes is orange-brown coloured, even though every node starts with a different value for that parameter. (2) The values of the different parameters in each node converge toward each other, as visible by the fact that all parameters eventually lose vividity and converge toward grey-brown colours. This can be characterised by the average standard deviation of all parameters of each node $\sigma_{ap}$. Note that, ignoring the effect of local training, $\sigma_{an}$ continuously drops at each round, i.e., the parameter sets of all nodes get closer and closer together. On the other hand $\sigma_{ap}$ does not go down indefinitely, meaning that the standard deviation of the set of parameters of each node converges to a positive value.}
    \label{fig:sigma-color-schematic}
\end{figure}

To understand the general characteristics of the learning process we propose a simplified numerical model: an iterative process, where each of the $n$ network nodes has a vector of $d$ parameters drawn from a Gaussian distribution with standard deviation $\sigma_\text{init.}$ and mean zero. At each iteration, each node updates its parameter vector by averaging its immediate neighbourhood, then, to emulate the effects of the local training step all node parameters are updated by adding a Gaussian distributed noise with zero mean and standard deviation $\sigma_\text{noise}$. Formally, using the notation introduced in Section~\ref{sec:notation}, for a given node $i$, its $j$-th parameter $w_{j,i}$ is initialized as $w_{j,i} \sim \mathcal{N}(0, \sigma_\text{init.})$ at communication round zero. At each subsequent iteration, the update rule is $w_{j,i} = \frac{1}{|\mathcal{N}_i|} \sum_{k \in \mathcal{N}_i} w_{j,k} + \mathcal{N}(0, \sigma_\text{noise})$. Here, the first term captures the parameter aggregation among neighbours, while the second term reflects the perturbation introduced by local training steps. This simplified model effectively approximates the behaviour of decentralised learning systems during early stages, as changes due to local training are typically negligible compared to the parameter adjustments from aggregation. Empirical evidence supporting this approximation can be seen in simulations of decentralised federated learning, as illustrated in \cref{fig:simulation_diffs}(a,b). A schematic visualisation of the aggregation process and the changes in $\sigma_{an}$ and $\sigma_{ap}$ is provided in \cref{fig:sigma-color-schematic}.

In general, this model shows many similarities with certain models of gossip protocol analysis \cite{boyd2005gossip} and the more simple DeGroot model of consensus and opinion dynamics \cite{degroot1974reaching}, with two major differences: First, unlike the model presented here, the original DeGroot model is a univariate model, though the possibility of a multi-variate version is briefly discussed and subsequently explored in other works \cite{ye2017analysis, parsegov2016novel}. Second, that the DeGroot and gossip protocol models do not involve the possible effects of the learning process that we modelled here using random noise.

The results from the simplified numeric model for random $k$-regular graphs predict, as shown in \cref{fig:simulation_diffs}(c), that the standard deviation of the value of the same parameter across nodes averaged for all parameters, which we call $\sigma_{an}$, will decrease to some value close to the standard deviation of noise (simulating changes due to local training). Meanwhile, the standard deviation of the parameters of the same node average across all nodes, $\sigma_{ap}$, will decrease only to a factor of $1/\sqrt{n}$ of the original standard deviation $\sigma_\text{init.}$. Note that for artificial neural networks each layer's initial parameters are usually drawn from distributions with different values of $\sigma_\text{init.}$, based on the number of inputs and outputs of each layer and other considerations. This does not prevent using the methodology presented here. The analysis here can be applied to each batch of parameters drawn from the same distribution, e.g., to the weights of each layer, independently. For details on the implementation of this, refer to \cref{alg:decavg}.

Two of the dynamics visualised in \cref{fig:simulation_diffs}(b,c) stand out in particular. First, the value towards which $\sigma_{ap}$ approaches can allow us to select the initial distribution of parameters $\sigma_\text{init.}$ in a way that after stabilisation of $\sigma_{ap}$, the neural network models would on expectation have an optimal parameter distribution. In \cref{sec:sigma_ap}, we show that the extent of compression of the parameters within each node, manifested as a reduction in $\sigma_\text{ap}$ can be calculated for any graph based on the distribution of eigenvector centralities of the nodes~\cite{newman2010Book}, with the case of graphs with uniform centralities giving a factor of $1/\sqrt{n}$. Second, the time to reach the steady state for $\sigma_{an}$ plays an important role since this determines the number of rounds required before the improvements of the learning process start in earnest. This is because the magnitude of the changes to parameters due to the learning process (modelled by noise in the numerical model) becomes comparable to those of the aggregation process. In \cref{sec:mixing-time} we show that this ``stabilisation'' time scales similarly, up to a constant factor, to the mixing time of lazy random walks on the graph.

\begin{algorithm}[hbt!]
\caption{Decentralised federated training cycle along with the\colortext{proposed initialisation steps.}}\label{alg:decavg}
\begin{algorithmic}[1]
\State{\textbf{given} number local batches $b$, Optimiser $Opt()$, set of neighbours $\mathcal{N}$.}
\State{\colorline{\textbf{estimate} $\left\| v_\text{steady} \right\|$ based on a sketch of the comms network, vide \cref{sec:estimating_v_steady}.}}
\For{layer $l$ with parameters $\omega^l_0$ in $w_0$ and activation function $f_l$}
    \State{\textbf{initialise} $\omega^l_0$ based on an method suitable for the architecture, e.g.~\citet{he2015delving}}
    \State{\colorline{\textbf{scale} $\omega^l_0$ by a factor of $\left\| v_\text{steady} \right\|^{-1}$}}
\EndFor{}
\Repeat
    \State{$t \leftarrow t + 1$}
    \State{$g_{t-1} \leftarrow \text{BatchGrads}(w_{t-1})$}
    \State{$w_t \leftarrow Opt(g_{t-1}, w_{t-1})$}
    \If{$t \mod b = 0$}
        \State{\textbf{send} parameters $w_{t}$ to neighbours}
        \State{\textbf{receive} parameters of all neighbours as $w^{i}_{t}\, \forall i \in \mathcal{N}$}
        \State{\textbf{aggregate} neighbourhood parameters as in \cref{eq:param-update}}\Comment{\textsc{DecAvg} aggregation}
        \State{\textbf{re-initialise} optimiser state.}
    \EndIf{}
\Until{\textit{stopping criteria are met}}
\end{algorithmic}
\end{algorithm}

\subsection{The compression of node parameters}\label{sec:sigma_ap}
For the simplified model we can analytically estimate the steady state values for $\sigma_{an}$ and $\sigma_{ap}$, as well as the scaling of the number of rounds to arrive at these values using methods from finite-state discrete-time Markov chains. Let $A$ be the adjacency matrix of our underlying graph $G$. We construct a right stochastic matrix $A'$ where
\begin{eqnarray}\label{eq:column-normal-markov-matrix}
    A'_{i j} = \frac{A_{i j} + I_{i j}}{\sum_k A_{k j} + I_{k j}}\,,
\end{eqnarray}
where $I$ is the identity matrix. This corresponds to the Markov transition matrix of random walks on graph $G$, if the random walker can stay at the same node or take one of the links connected to that node with equal probability for each possible action. This formulation can also be seen on the basis of the \textsc{DecAvg} aggregation in \cref{eq:param-update}. If we arrange all initial node parameters in a $d \times n$ matrix $W_\text{init.}$, the parameters at round $t$ are determined by $W_\text{init.} A'^t + \sum^{t - 1}_{\tau = 0} N_\tau A'^\tau$, where $N_\tau$ is a random $d \times n$ noise matrix with each index drawn from $\mathcal{N}(0, \sigma_\text{noise}^2)$. 

Assuming that the graph $G$ is connected, the matrix $A'^t$ would converge to a matrix where each row is the steady state vector of the Markov matrix $A'$, the eigenvector corresponding to the largest eigenvalue 1, normalised to sum to 1. If the steady state vector is given as $v_\text{steady}$ the variance contribution of the term $W_\text{init.} A'^t$ along each row (i.e., the expected variance of parameters of each node) is given as $\sigma_\text{init.}^2 \left\| v_\text{steady} \right\|^2$. It can be trivially shown, from a direct application of the Cauchy--Schwarz inequality $|\langle \vec{1}, v_\text{steady} \rangle| \leq \|\vec{1}\|  \|v_\text{steady}\|\,$ that for connected networks $\langle \vec{1}, v_\text{steady} \rangle = \sum v_\text{steady} = 1$, the $\left\| v_\text{steady} \right\|^2$ term has a minimum value of $1/n$. This is achieved for random regular networks and other network models where nodes have uniformly distributed eigenvector centralities\footnote{Eigenvector centrality is a node characteristic ~\cite{newman2010Book}) calculated from the eigenvector equation $Ax=\lambda_{\text{max}} x$, where $\lambda_{\text{max}}$ is the greatest eigenvalue of $A$. The eigenvector centrality of node $i$ is $x_i/\sum_j x_j$. Here the eigenvector centrality as calculated from $A'$ is meant.}, such as Erdős–Rényi networks and lattices on $d$-dimensional tori.

Given that the noises are drawn independently, the variance contribution of the noise term $\sum_{\tau=0}^{t-1} N_\tau A'^\tau$ has an upper-bound of $t \sigma^2_\text{noise}$. If $t \sigma^2_\text{noise} \ll \sigma^2_\text{init}/n$, then the standard deviation across parameters at round $t$ can be approximated by
$ \lim_{t \rightarrow \infty} \sigma_{ap} \approx \sigma_\text{init.} \left\| v_\text{steady} \right\|$. For a large connected random $k$-regular network, this reduces to $\lim_{t \rightarrow \infty} \sigma_{ap} = \sigma_\text{init.}/\sqrt{n}$.
Generally for other networks, numerical solutions for $\left\| v_\text{steady} \right\|$ can be obtained by calculating sum-normalised eigenvector centralities of the original network after adding self-loops to all nodes, with weights equal to the node degree.

These results, combined with the existing analyses on the role of artificial neural network parameters and their effect on diminishing or exploding gradients~\cite{he2015delving} suggest that it is reasonable to take into account the compression of the node parameters (e.g., the $1/\sqrt{n}$ factor for $k$-regular networks) when initialising the parameters, and we show in \cref{sec:estimating_v_steady} that even a rough estimate of this factor would be quite effective in practice. Depending on the choice of architecture and optimiser, and especially for large networks with hundreds of nodes, this re-scaling of initial parameters can play a sizeable role in the efficacy of the training process. In cases where we were dealing with random $k$-regular or Erdős–Rényi networks, as is the case for the majority of the experiments in this paper, we took this into account by simply multiplying a gain factor of $\sqrt{n}$ in the standard deviation of layer parameters suggested by \citet{he2015delving}. 

\begin{tcolorbox}[colback=gray!10!white, colframe=black, title=Proposed Initialization Strategy]
Let $\sigma_{\text{init}}$ be the standard deviation of layer parameters as suggested by~\citet{he2015delving}. In decentralised settings, we propose the following adjusted initialization:
\begin{itemize}
    \item \textbf{Exact:} Use $\sigma_{\text{init}} \cdot \left\| v_\text{steady} \right\|^{-1}$, where $\left\| v_\text{steady} \right\|$ is the $\ell_2$-norm of the steady-state eigenvector (corresponding to eigenvalue 1) of the Markov matrix $A'$ (Equation~\eqref{eq:column-normal-markov-matrix}) associated with the communication graph $G$, normalized to have unit sum.

    \item \textbf{Approximate:} Use $\sigma_{\text{init}} \cdot \left\| v_\text{steady} \right\|_{approximate}^{-1}$, where $\left\| v_\text{steady} \right\|_{approximate}$ is obtained as described in Section~\ref{sec:estimating_v_steady}. 
\end{itemize}

In practice, this desired correction to He's standard deviation can be achieved by multiplying the weights initialized according to~\citet{he2015delving} by the inverse of $\left\| v_\text{steady} \right\|$ (exactly or approximately computed). This follows from the scaling properties of standard deviation ($\sigma(aX) = a\sigma(X)$, where $X$ is a generic random variable and $a$ is a constant value). The complete pipeline integrated with the learning steps is sketched in \cref{alg:decavg}.
\end{tcolorbox}

Note that $v_\text{steady}$ is simply the sum-normalised vector of eigenvector centralities of the communication network nodes with a self-loop added to all nodes with a weight equal to the degree of that node. Each element of that vector specifies the probability of a random walk to end up on that specific node, if the random walk process has equal probability of taking any one of the edges or staying on the node. This means that the value of this gain is a factor of the system size and the distribution of network centralities. The practical application of this is explored in \cref{sec:estimating_v_steady}.

The numeric model applies with minimal changes to directed and weighted communication networks, similar to connected undirected networks. In the case of a strongly connected directed network, the convergence is guaranteed since the stochastic matrix $A'$ is aperiodic due to the existence of the self-loops. For the case of a weighted communication network, the weights are reflected in the graph adjacency matrix $A$, with the provision that a diagonal matrix of the weights each node assigns to its own weights should be used in \cref{eq:column-normal-markov-matrix} instead of the identity matrix $I$.

\subsection{Estimating parameter scaling factor \texorpdfstring{$\left\| v_\text{steady} \right\|$}{||v\_steady||}}\label{sec:estimating_v_steady}
Calculating the scaling factor $\left\| v_\text{steady} \right\|$ based on a perfect knowledge of the entire communication network is trivial. In real-world  scenarios, however, it is often the case that we can only rely on each node's imperfect knowledge of the connectivity network during initialisation. In this section, we explore a few scenarios to illustrate how this affects the process described.

Often, while the full topology of the communication network might not be known to each node, the network might have emerged as a result of a central organising principle. For example, assume that the communication network is formed through a peer discovery system where a new node is assigned to be connected to existing nodes. If the assignment probability is a linear function of their current degree (i.e., a node with more neighbours is more likely to be recommended to a new node) then the resulting network would be an example of the preferential attachment process, with a power-law degree sequence \cite{newman2005power}. Similarly, if the communication networks are based on real-world social relationships or the physical distance of the nodes, then the network would have properties similar to those observed in social or geometric networks, respectively. If such information about the network formation principles is known beforehand, fewer variables need to be estimated.

\begin{figure}
    \centering
    \includegraphics[width=0.70\linewidth]{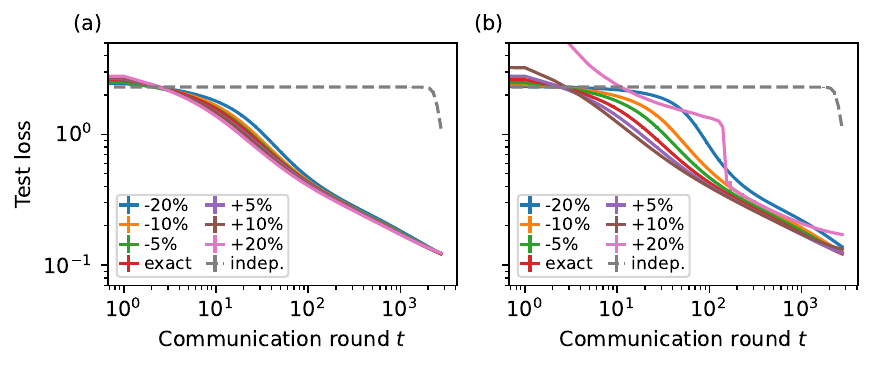}
    \caption{The effect of incomplete information in form of over- or under-estimating (a) the number of nodes or (b) the $\left\| v_\text{steady} \right\|$ size exponent (vide \cref{fig:toy-model-gain-scaling}(a,b)) on our proposed initialisation method. Note that our proposed initialisation still performs significantly better than unscaled independent initialisation, shown with dashed lines. Performed on the MLP configuration with MNIST dataset, on fully connected network. Error bars represent 95\% confidence intervals}
    \label{fig:estimates}
\end{figure}

Let us take, for example, a communication network formed based on randomly establishing connections between two nodes, or formed one based on a Barabási--Albert preferential attachment process. In these cases, with only an estimate of the number of nodes $n$, it is possible to estimate the scaling factor $\left\| v_\text{steady} \right\|$. While this information is not necessarily available to all nodes, a simple application of a gossip protocol \cite{boyd2005gossip} can provide an estimate to all nodes. It is important to note that the estimate for $n$ need not be exact. \cref{fig:estimates}(a) shows that even in the case of a substantial over- or under-estimation of the number of nodes, our proposed initialisation method still performs quite close to when it is presented with perfect knowledge of the network.

\begin{figure}[!bth]
    \centering
    \includegraphics[width=0.98\linewidth]{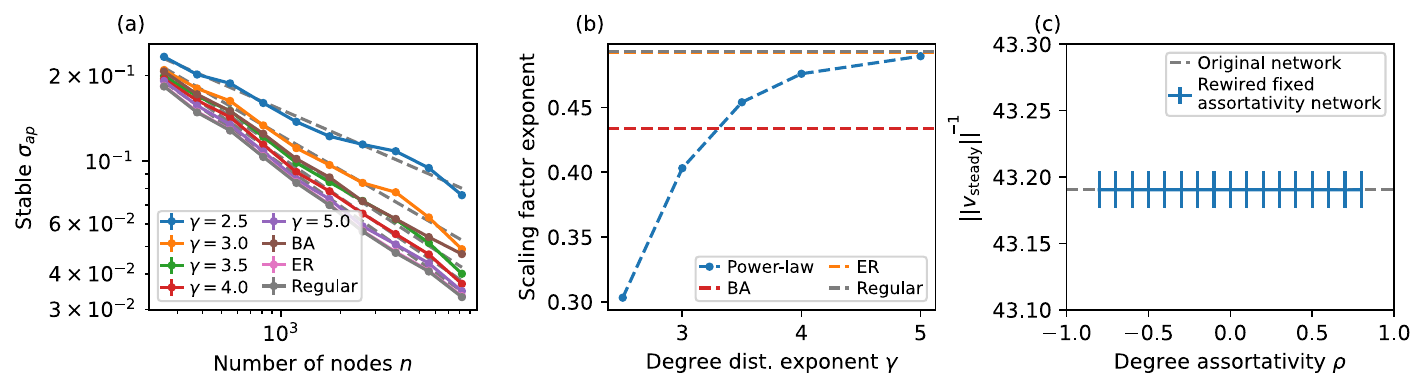}
    \caption{(a,b) The effect of heterogeneous distribution of centralities in the scaling factor $\left\| v_\text{steady} \right\|$ from the simplified numerical model. Homogeneous random networks (Erdős--Rényi $G(n, p)$ networks and random $k$-regular networks) display $\left\| v_\text{steady} \right\| \approx 1/\sqrt{n}$, while Barabási--Albert networks and configuration model heavy-tail degree distribution networks (with degree distribution $p(k)\sim k^{-\gamma}$) show this factor scaling exponentially with the number of nodes with different exponents that is itself a function of $\gamma$. (c) Degree distribution preserving rewiring of Erdős--Rényi network ($n=2048$) to produce networks with various values of degree assortativity $\rho$ shows that $\left\| v_\text{steady} \right\|$ is not affected by degree assortativity. Error bars represent 95\% confidence intervals.}
    \label{fig:toy-model-gain-scaling}
\end{figure}

If no information on the network topology is known in advance, it is possible to arrive at a best guess by polling a sample of the network (perhaps through a gossip protocol) for a degree distribution. The scaling of $\left\| v_\text{steady} \right\|$ with system size in random networks with different distributions is illustrated in \cref{fig:toy-model-gain-scaling} where the value of the scaling factor exponent is derived for Erdős--Rényi, $k$-regular, Barabási--Albert and heavy-tail degree distribution configuration model random networks with the same size and (on expectation) the same number of links. The value of $\left\| v_\text{steady} \right\|$ is simply $n^{-\alpha}$ where $\alpha$ is the scaling factor exponent in \cref{fig:toy-model-gain-scaling}(a,b). 

Furthermore, we show empirically that $\left\| v_\text{steady} \right\|$ is independent of the degree assortativity of the network. This is done by rewiring a network using the edge swap method, i.e.~selecting two edges and swapping the endpoints. This is performed through simulated annealing: a specific target value for degree-assortativity is set and random edge-swaps are accepted or rejected based on their utility and a temperature variable that decreases slowly over time, until the network converges to the desired target assortativity. The results in \cref{fig:toy-model-gain-scaling}(c) show that $\left\| v_\text{steady} \right\|$ stays the same after rewiring. 

\subsection{Initial stabilisation time}\label{sec:mixing-time}
The stabilisation time of $\sigma_{an}$, the number of communication rounds until the blue curve in \cref{fig:simulation_diffs}(b,c) flattens out, determines the number of rounds where local training has a negligible effect on the parameters. Understanding the scaling of stabilisation time with the number of nodes and other environmental parameters is important, as before this stabilisation the aggregation process dominates the local training process by several orders of magnitude (\cref{fig:simulation_diffs}(a)), inhibiting effective training.

Deriving how the number of rounds until stabilisation of $\sigma_{an}$ scales with the number of nodes $n$ is a matter of calculating the mixing time of the Markov matrix $A'$. The problem is remarkably close to the lazy random walk, where at each step the walker might stay with probability $1/2$ or select one of the links for the next transition. However, in our case, this staying probability is lower than or equal to that of a lazy random walk, being equal to $1/(k_i + 1)$ where $k_i$ is the degree of node $i$. It has been shown that, since the staying probabilities are bounded in $(0, 1)$, the mixing time of the random walk process described here grows asymptotically with that of lazy random walk up to a constant factor~\citep[Corollary 9.5]{peres2015mixing}.

The mixing time of lazy random walks on graphs is a subject of active study. Lattices on $d$-dimensional tori have a mixing time with an upper bound at $d^2 l^2$~\citep[Theorem 5.5]{levin2017markov} where $l \propto n^{1/d}$ is the linear system size. Connected random $k$-regular networks, as expander graphs, have a mixing time of $O(\log n)$~\cite{barzdin1993realization, pinsker1973complexity}, while connected supercritical Erdős–Rényi ($G(n, m)$ and $G(n, p)$) graphs (with average degree larger than 1) have lazy random walk mixing times of $O(\log^2 n)$~\cite{fountoulakis2008evolution, benjamini2014mixing}. Generally, for a given Markov matrix the convergent rate can be estimated based on its spectral gap~\cite{mufa1996estimation}.

\section{Scalability and role of learning parameters}\label{apdx:topology}
% Make the link very clear
As shown before in \cref{fig:simulation_plateau}, the choice of initialisation strategy significantly affects the behaviour of the system when varying the environment parameter such as the number of nodes. In this section, we will briefly discuss the effect of network topology on the learning trajectory of the system, then systematically analyse the role of different environmental parameters such as the system size (number of nodes), the communication network density, the training sample size and the frequency of communication between nodes in the trajectory of the decentralised federated learning, when using the initialisation method proposed in \cref{sec:initialisation}. As most of these quantities are involved in some form of cost-benefit trade-off, understanding the changes in behaviour due to each one can allow a better grasp of the system behaviour at larger scales.

For the rest of this section, however, we limited the analysis to a single topology, random $k$-regular networks, to focus on a more in-depth analysis of the role of environmental parameters other than the network topology, such as the system size, frequency of communication, and network density.

For the purposes of this section, we make the simplifying assumption that the communication time is negligible compared to the training time. In some cases, we introduce ``wall-clock equivalent'' values, indicating the computation time spent by an individual node up to communication round $t$, multiplied by the number of training mini-batches of training between two rounds of communication. This ``wall-clock equivalent'' can be seen as a linear scaling of the communication rounds $t$.

\begin{figure}[!tb]
    \centering
    \includegraphics[width=0.7\linewidth]{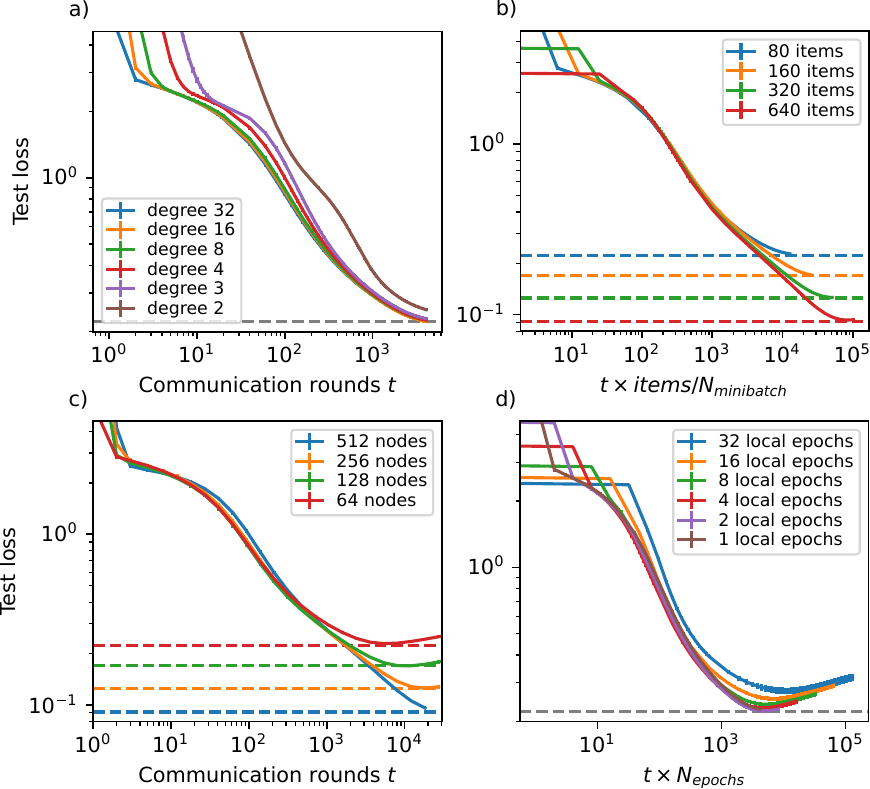}
    \caption{Trajectory of mean test cross-entropy loss over communication rounds for (a) connected random $k$-regular networks with $n=64$ nodes and different values for degree $k$, with 80 balanced training samples per node, (b) 32-regular random network with different number of total labelled training samples, balanced across classes, assigned to each item, (c) with different number of nodes and (d) with different number of local epochs between communications. In all panels, the horizontal dashed lines correspond to the best test loss of a central system with the same number of total training samples as the entire decentralised federated learning system simulated. Error bars represent 95\% confidence intervals. The horizontal axes in (b,d) are scaled to show the ``wall-clock equivalent'', a value linearly comparable to the total computation cost of a single node until round $t$.
    }
    \label{fig:simulation_params}
\end{figure}

\paragraph{Network density}
The number of links in the communication network directly increases the communication burden on the nodes. Our results (\cref{fig:simulation_params}(a) show that while a very small value for the average degree affects the rapidity of the training convergence disproportionately, as long as the average degree is significantly larger than the critical threshold for connectivity, i.e., for random network models with average excess degree $\langle q(k) \rangle \gg 1$, the trajectory will be quite consistent across different network densities. Note that, although in \cref{sec:mixing-time} we were mostly concerned with the scaling of the initial mixing time with the number of nodes, in many cases this would also benefit from a higher average degree. Also note that average degrees close to the critical threshold might not prove practical or desirable for the communication network in the first place, as the network close to the critical threshold is highly susceptible to fragmentation with the cutting off of even very few links. 

\paragraph{Training samples per node}
Assuming that each device is capable of performing training on a constant rate of mini-batches per unit of time, more training samples per node increase the total amount of training data, while also linearly increasing the training time for every epoch. Our results (\cref{fig:simulation_params}(b)) show that (1) the test loss approaches that of a centralised system with the same number of total training samples, and (2) that the trajectory of test loss with effective wall-clock time remains consistent.

\paragraph{System size and total computation cost}
The number of nodes in the network affects the training process in multiple ways. Suppose a larger system size is synonymous with a proportionate increase in the total number of training samples available to the system as a whole. In that case, it is interesting to see if the system is capable of utilising those in the same way as an increase in the number of items per node would. Our results (\cref{fig:simulation_params}(c)) show that if the increase in size coincides with an increase in the total number of items, the system is able to effectively utilise these, always approaching the test loss limit of a centralised system with the same total data.

\begin{figure}[!hbt]
    \centering
    \includegraphics[width=0.42\linewidth]{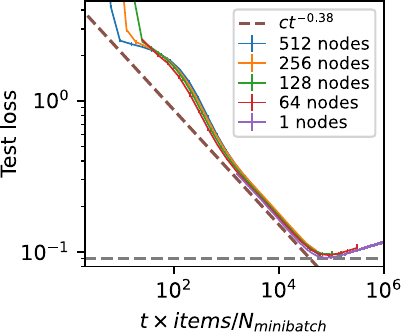}
    \caption{%$(a)
    Trajectory of mean test cross-entropy loss over wall-clock time equivalent over 32-regular random graphs and for an isolated node while keeping the total number of training samples across the whole system constant. Each node was assigned training samples balanced across 10 classes, with a total of 40\,960 training samples divided equally across the nodes. Error bars represent 95\% confidence intervals. The horizontal dashed line corresponds to the best test loss of a central system with the same total amount of training samples as the entire decentralised federated learning system simulated. The sloped dashed line shows the power-law trajectory of loss with equivalent to wall-clock time consistent with results from \citet{henighan2020scaling}.
    }
    \label{fig:simulation_speedup}
\end{figure}

Another aspect is that an increase in the number of nodes would mean an increase in the total computation cost, so it would be interesting to analyse if this increase (without a corresponding increase in the total amount of data) would result in any improvements in the learning trajectory. In short, our results in \cref{fig:simulation_speedup} show that if the same amount of data is spread across more nodes, each node will have to train on roughly the same number of minibatches to arrive at a similar test loss, and that this result is even consistent with the learning trajectory of the centralised single node scenario.

\paragraph{Communication frequency}
Finally, we consider the role of communication frequency in the trajectory of loss, measured by the number of local training epochs between communications. It has been shown in the context of decentralised parallel stochastic gradient descent that a higher frequency of communications increases the efficacy of the training process, as it prevents a larger drift~\cite{lian2017can}. While a similar phenomenon in the context of an uncoordinated decentralised federated learning seems plausible, showing this relationship empirically on a system of reasonable size was fraught with difficulties due to the issues discussed in \cref{fig:simulation_plateau}. Utilising the proposed initialisation method enables this and allows us to confirm (\cref{fig:simulation_params}(d)) that while more frequent communication increases the communication burden on the entire network, more frequent communication translates to both a lower final test loss as well as faster convergence.

\section{Conclusion}
Here we introduced a fully uncoordinated, decentralised artificial neural network initialisation method that provides a significantly improved training trajectory, while solely relying on the macroscopic properties of the communication network. We also showed that the initial stages of the uncoordinated decentralised federated learning process are governed by dynamics similar to those of the lazy random walk on graphs. Furthermore, we also showed empirically (\cref{apdx:topology}) that when using the proposed initialisation method, the test loss of the decentralised federated learning system can approach that of a centralised system with the same total number of training samples, and that the final outcome, in terms of the best test loss achieved, is fairly robust to different network densities and momentary communication failures, and it can benefit from more frequent communication between the nodes.

The proposed initialisation method works as an extension to any existing neural network initialisation methods. While in this work we used the example of the initialisation method proposed by \citet{he2015delving}, when the correction factor $\left\| v_\text{steady} \right\|$ is applied to the initial parameters generated through any initialisation method, it ensures that after an initial relaxation phase the parameters are not unduly compressed due to the effects of successive averaging in each aggregation step.

\subsection{Limitations}\label{sec:limitations}
In this work, we have not considered an unequal allocation of computation power among the nodes to focus solely on the role of the initialisation and the network. 
In real-world settings, these are often combined or correlated with network properties such as the degree or other centrality measures, which might affect the efficacy of the decentralised federated learning process. Understanding the combination and interactions of these properties with network features adds another layer of interdependency and complexity to the problem, which most certainly was not addressable without first studying the simpler case presented here. The prospect of extending this work to these more complex settings is interesting to consider.

While in this paper we discussed the \textsc{DecAvg} aggregation, perhaps one of the most commonly studied aggregation strategies in the context of federated learning, the same approach can be trivially extended to any aggregation method that relies on a linear combination of parameters of the entire neighbourhood where all multipliers are non-zero.

The federated learning process presented here does not support heterogeneous machine learning architectures between nodes. We expect this to become more prominent with the advances in edge computing and device availability. We also did not consider possible heterogeneities in node-to-node communication patterns, such as burstiness or diurnal pattern, which have been shown to affect the rapidity of other network dynamics like spreading and percolation processes~\cite{karsai2011small, badie2022directed_long, badie2022directed_short}.

Additionally, some artificial neural network architectures utilise batch normalisation~\cite{ioffe2015batch}, which seemingly greatly limits (but does not eliminate \cite{he2015delving}) the vanishing/exploding gradients issue that necessitates careful parameter initialisation. It is important to note that this greatly reduces options in the choice of architecture or risks introducing gradient explosion at initial training steps, making deep networks of arbitrary structure prohibitively difficult to train~\cite{yang2019mean}. Careful parameter initialisation, on the other hand, provides a more generalisable solution.

Also of note is that this work only covers the simplest forms of temporal dynamics, namely activating or deactivating communication links and isolating entire nodes through a Poisson process. While a more in-depth temporal analysis is outside the scope of this work, in a future manuscript we will provide a more extensive study of the role of temporal heterogeneities on the dynamics of the decentralised federated learning process.

This work enables uncoordinated decentralised federated learning that can efficiently train a model using all the data available to all nodes without having the nodes share data directly with a centralised server or with each other. While this enables or streamlines novel use cases, it is important to note that trained machine learning models themselves could be exploited to extract information about the training data~\cite{carlini2021extracting, carlini2023extracting2}. It is therefore important not to view federated learning as a panacea for data privacy issues, but to view direct data sharing as the weakest link in data privacy.

\section*{Declarations}

\subsection*{Availability of data and materials}
All data generated or analysed during this study are included in this published article. Complete information on access to datasets and experimental/simulation codes as well as reproduction instructions are provided in \cref{apdx:data-and-experiments}. All datasets and codes used in this work are available under permissive licenses and are publicly archived.

\subsection*{Funding}
This research was supported by CHIST-ERA-19-XAI010 SAI projects, FWF (grant No. I 5205). JK acknowledges partial support by ERC grant No. 810115-DYNASNET. MK acknowledges support from the ANR project DATAREDUX (ANR-19-CE46-0008); the SoBigData++ H2020-871042 project; and the National Laboratory for Health Security, Alfréd Rényi Institute, RRF-2.3.1-21-2022-00006. CB's work was partly funded by the PNRR - M4C2 - Investimento 1.3, Partenariato Esteso PE00000013 - ``FAIR'', LV's work was partially supported by the European Union - Next Generation EU under the Italian National Recovery and Resilience Plan (NRRP), Mission 4, Component 2, Investment 1.3,CUP B53C22003970001, partnership on ``Telecommunications of the Future'' (PE00000001 - program ``RESTART''). STRIVE - Sciences for Industrial, Green and Energy Transitions - MeSAS subproject - Models and Tools for Sustainable AI.

\subsection*{Authors' contributions}
All authors participated in conceptualisation, developing the methodology and review and editing of the manuscript. ABM wrote the software, conducted the experiments and validations and wrote the original draft of the manuscript.

\subsection*{Acknowledgements}
We acknowledge the EuroHPC Joint Undertaking for awarding this project access to the EuroHPC supercomputer LUMI, hosted by CSC (Finland) and the LUMI consortium through a EuroHPC Regular Access call. We also acknowledge the computational resources provided by the Aalto Science-IT project.

\bibliography{references}

% \newpage
% \appendix

\begin{appendices}

\section{Datasets, implementation and experimental architecture}\label{apdx:data-and-experiments}

The artificial neural network architectures employed in this manuscript are a simple feedforward neural network with four fully connected layers, consisting of 512, 256, and 128 neurons in three hidden layers, followed by an output layer of size 10, and employs \textit{ReLU} activation functions after each layer except the output layer, a simple feedforward neural network consisting of 3 2D convolutional layers with 32, 64 and 64 output channels, each with 3 kernels and one pixel padding of zeros, followed by two fully connected linear hidden layers of size 128 and 64 and one output layer and the VGG16 architecture~\cite{simonyan2014very}. Of course, as discussed in the literature, the effects of the initialisation method would be even more visible in deeper neural network architectures~\cite{he2015delving, glorot2010understanding}.

Our experiments will be performed on subsets of the MNIST digit classification task~\cite{lecun1998gradient}, the So2Sat LCZ42 dataset for local climate zone classification~\cite{zhu2019so2sat} and the CIFAR-10 image classification dataset~\cite{krizhevsky2009learning}, distributed between nodes either iid or non-iid based on a Zipf distribution. In terms of local optimisation, we tested stochastic gradient decent with momentum and Adam with decoupled weight decay, although empirical evidence (\cref{fig:simulation_diffs}) hints that the the effects of local optimisation are only non-negligible compared to those of the aggregation at a longer time-scale than the one mainly of interest in this manuscript.

The MNIST dataset was released under the MIT license. Available at \url{https://huggingface.co/datasets/ylecun/mnist}.

The So2Sat LCZ42 dataset was released under the Creative Commons 4.0 Attribution licence, at \url{https://mediatum.ub.tum.de/1613658}. In our use case, we used a random subset of the random split of the third version, only including the 10 bands from the Sentinel-2 satellite to artificially simulate a more realistic, data-poor scenario.

The CIFAR-10 (Canadian Institute for Advanced Research, 10 classes) dataset was released under the Creative Commons 0 version 1.0, available at \url{https://www.cs.toronto.edu/~kriz/cifar.html}.

\cref{tab:configs} shows in brief the information about the configurations used for producing figures in the manuscript.

\begin{table}[!htb]
\centering
\caption{Configurations used in this manuscript. Stochastic gradient descent used momentum $m=0.5$, while  Adam optimiser with decoupled weight decay was initialised with parameter $\beta_1 = 0.9$, $\beta_2 = 0.999$, $\epsilon=10^{-8}$ and $\lambda=10^{-2}$. Both Optimisers used a learning rate of $10^{-3}$. All configurations used a minibatch size of 16 and 8 minibatches of local training per communication round.}
\label{tab:configs}
\begin{tabular}{@{}llllllr@{}}
\toprule
Cfg. & Dataset & Architecture & Comm. net. & Optimiser & Data dist. & \multicolumn{1}{l}{Items per node} \\ \midrule
A & MNIST    & MLP     & Full      & SGD   & iid              & 512  \\
B & So2Sat   & CNN+MLP & BA (m=8)  & SGD   & Zipf ($\alpha=1.8$) & 1024 \\
C & CIFAR-10 & VGG-16  & 4-regular & SGD   & iid              & 512  \\
D & MNIST    & MLP     & Full      & AdamW & iid              & 512  \\ \bottomrule
\end{tabular}
\end{table}

Runtimes for and configuration used is reported in \cref{tab:runtimes} for the purpose of reproduction.

The implementation of the full-fidelity simulated decentralised federated learning system is available for the purposes of reproduction under the MIT open-source license at \url{https://github.com/arashbm/sat}. The development of this works relied on various pieces of open-source scientific software \cite{paszke2019pytorch,badie2023reticula,tange2023gnu,harris2020array,beranek2024hyperqueue}.

\begin{table}[!htb]
\centering
\caption{Median runtime (in minutes) for a single realisation of each configuration, and the total runtime of all realisations for each configuration. Scaling refers to \cref{fig:simulation_plateau}, Estimates to \cref{fig:estimates}, and Probs to \cref{fig:occupation}. Runtimes were measured on AMD MI250X GPUs with two realisations running concurrently per graphics compute die (GCD) to increase efficiency. Configuration labels refer to \cref{tab:configs}. Note that if the preliminary experiments and the experiments presented solely in the appendices are taken into account, the total computation costs increases to roughly 10000--12000 GCD-hours.}
\label{tab:runtimes}
\begin{tabular}{@{}lrrr@{}}
\toprule
Configuration & \multicolumn{1}{l}{Size (nodes)} & \multicolumn{1}{l}{Median runtime (mins)} & \multicolumn{1}{l}{Total runtime (hours)} \\ \midrule
A (Scaling)     & 8                    & 9.3                  & 10.1     \\
                & 16                   & 20.1                 & 22.4     \\
                & 32                   & 51.2                 & 55.1     \\
                & 64                   & 137.1                & 148.3    \\ \midrule
A (Estimates)   & 64                   & 136.2                & 642.4    \\ \midrule
A (Probs)       & 64                   & 77.4                 & 1466     \\ \midrule
B (Scaling)     & 32                   & 95.0                 & 405.2    \\ \midrule
                & 64                   & 177.7                & 757.7    \\
                & 128                  & 340.2                & 1499.4   \\
                & 256                  & 716.8                & 1530.1   \\ \midrule
C (Scaling)     & 8                    & 47.5                 & 25.7     \\
                & 16                   & 92.9                 & 49.7     \\
                & 32                   & 167.1                & 90.0     \\
                & 64                   & 314.9                & 169.87   \\ \midrule
D (Scaling)     & 8                    & 10.54                & 5.5      \\
                & 16                   & 23.05                & 12.0     \\
                & 32                   & 51.42                & 27.3     \\
                & 64                   & 126.8                & 68.2     \\ \midrule
Total runtime   & \multicolumn{1}{l}{} & \multicolumn{1}{l}{} & 6985.0  \\
Total GCD-hours & \multicolumn{1}{l}{} & \multicolumn{1}{l}{} & 3492.5
\end{tabular}
\end{table}

\end{appendices}

\end{document}